\documentclass[11pt]{article}

\usepackage{emnlp2022}

\newcommand\blfootnote[1]{%
  \begingroup
  \renewcommand\thefootnote{}\footnote{#1}%
  \addtocounter{footnote}{-1}%
  \endgroup
}

\usepackage{times}
\usepackage{latexsym}
\usepackage{graphicx}
\usepackage{todonotes}
\usepackage{comment}
\usepackage{amsmath}
\usepackage{enumitem}
\usepackage{soul,xcolor}
\usepackage{arydshln}
\usepackage{multirow}
\usepackage{booktabs}
\usepackage{mathtools, nccmath}
\usepackage{float}
\usepackage{amsfonts} 
\usepackage{amsmath}

\usepackage[T1]{fontenc}

\usepackage[utf8]{inputenc}

\usepackage{microtype}

\usepackage{tikz}
\usetikzlibrary{shapes,snakes,positioning,backgrounds,calc}

\newcommand{\dataset}{\texttt{CURT}}

\newcommand{\model}{\texttt{DABERTa}}

\usepackage{multirow}
\usepackage{colortbl}

\usepackage{xcolor, soul}

\definecolor{mintgreen}{rgb}{0.6, 1.0, 0.6}
\definecolor{cherryblossompink}{rgb}{1.0, 0.72, 0.77}
\DeclareRobustCommand{\hlcorrect}[1]{{\sethlcolor{mintgreen}\hl{#1}}}
\DeclareRobustCommand{\hlwrong}[1]{{\sethlcolor{cherryblossompink}\hl{#1}}}

\usepackage{multirow}
\usepackage{enumitem}

\usepackage{hyperref}
\hypersetup{
  colorlinks   = true,    
  urlcolor     = darkblue,    
  linkcolor    = darkblue,    
}

\begin{document}
\title{\textit{Empowering the Fact-checkers!} Automatic Identification\\ of Claim Spans on Twitter}

\author{Megha Sundriyal$^1$$^{*}$, Atharva Kulkarni$^1$$^{*}$, Vaibhav Pulastya$^1$, Md Shad Akhtar$^1$, Tanmoy Chakraborty$^2$\\
        $^1${IIIT Delhi, India}, $^2${IIT Delhi, India} \\
        \textit{\{meghas, atharvak, vaibhav17271, shad.akhtar\}@iiitd.ac.in},  \textit{tanchak@ee.iitd.ac.in}
         }
        


\maketitle

\begin{abstract}
The widespread diffusion of medical and political claims in the wake of COVID-19 has led to a voluminous rise in misinformation and fake news. The current vogue is to employ manual fact-checkers to efficiently classify and verify such data to combat this avalanche of claim-ridden misinformation. However, the rate of information dissemination is such that it vastly outpaces the fact-checkers' strength. Therefore, to aid manual fact-checkers in eliminating the superfluous content, it becomes imperative to automatically identify and extract the snippets of claim-worthy (\textit{mis})\textit{information} present in a post. In this work, we introduce the novel task of \textit{Claim Span Identification} (\textit{CSI}). We propose  \textbf{\dataset}, a large-scale Twitter corpus with token-level claim spans on more than $7.5k$ tweets. Furthermore, along with the standard token classification baselines, we benchmark our dataset with \textbf{\model}, an adapter-based variation of RoBERTa. 
The experimental results attest that \model\ outperforms the baseline systems across several evaluation metrics, improving by about $1.5$ points. We also report detailed error analysis to validate the model's performance along with the ablation studies. Lastly, we release our comprehensive span annotation guidelines for public use.
\end{abstract}

\section{Introduction}
\label{sec:intro}
\begin{figure}
    \includegraphics[width=0.45\textwidth]{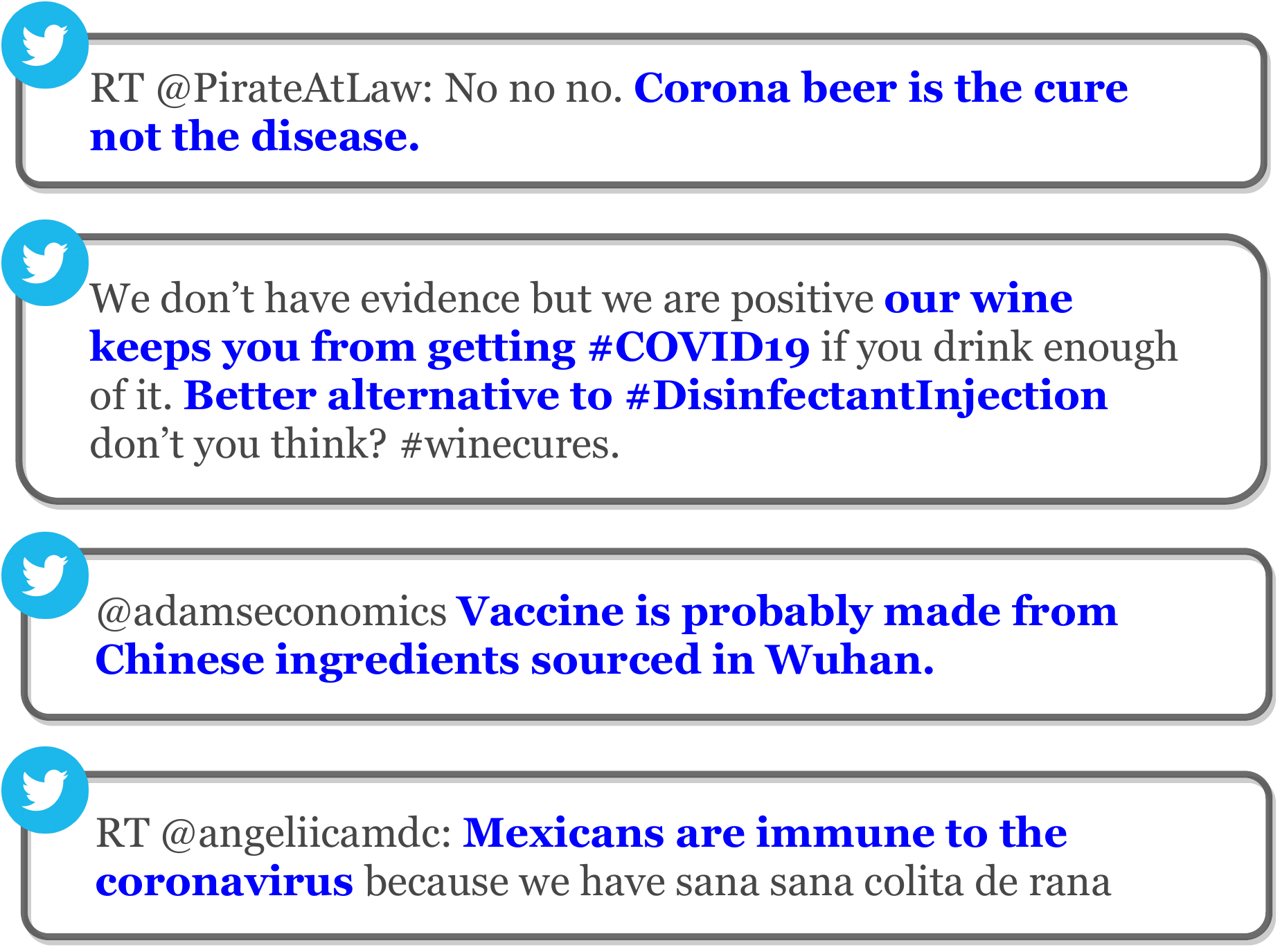}
    \vspace{-0.5em}
    \caption{Examples  of  claim  tweets  and  their  ground truth claim spans highlighted in boldface text (blue).}
    \label{fig:span-examples}
    \vspace{-1.5em}
\end{figure}

\blfootnote{$^*$Equal contribution}The swift acceleration of Online Social Media (OSM) platforms has led to tremendous democratized content creation and information exchange. Consequently, these platforms serve as ideal breeding grounds for malicious rumormongers and talebearers, abetting a colossal upsurge of misinformation. Such misinformation manifests in many ways, including bogus claims, fabricated information, and rumors. The massive COVID-19 \textit{ `Infodemic'} \citep{naeem2020covid} is one such malignant byproduct that led to the rampant spread of political and social calumny \citep{ferrara2020types, margolin2020fbi, ziems2020racism}, accompanied by counterfeit pharmaceutical claims \citep{o2020going}. Therefore, finding such claim-ridden posts on OSM platforms, investigating their plausibility, and differentiating the credible claims from the apocryphal ones has risen to be a pertinent research problem in Argument Mining (AM). 

\textit{`Claim'}, as coined by \citet{toulmin1958uses}, is \textit{`an assertion that deserves our attention'}. It is the key component of any argument \citep{daxenberger-etal-2017-essence}. Consider the second tweet, \textit{`We don't have evidence...'}, as given in Figure~\ref{fig:span-examples}. For the task of claim identification at the coarse level, the entire tweet will be marked as a claim. However, on closer inspection, we find that the text fragments of \textit{`our wine keeps you from getting \#COVID19'} and \textit{`Better alternative to \#DisinfectantInjection'} represent the finer argumentative units of claim and form the set of evidence, based on which this tweet is considered a claim. Segregating such argumentative units of misinformed claims from their benign counterparts fosters many benefits. To begin with, it partitions the otherwise independent claims in a single post, enabling us to retrieve a larger number of claims. Secondly, it acts as a precursor to the downstream tasks of claim check-worthiness and claim verification. Thirdly, it will also bring in the angle of {\em explainability} in coarse-grained claim identification. Finally, it will serve the manual fact-checkers and hoax-debunkers\footnote{\url{https://www.snopes.com/}}$^{,}$\footnote{\url{ https://www.politifact.com/}} to conveniently strain out the unnecessary shreds of text from further processing. We further elaborate on the necessity of claim span identification and exemplify it in Section~\ref{sec:motivation}.

Though the recent literature reflects extensive work done in claim detection \citep{daxenberger-etal-2017-essence, chakrabarty2019imho, gupta-etal-2021-lesa}, limited forays have been made in claim span identification i.e., recognizing the argumentative components of a claim \citep{wuhrl2021claim}. In the recent past, commendable work has been done on span-level argument unit recognition pertaining to other computational counterparts under the umbrella of AM, such as hate speech \citep{mathew2020hatexplain}, toxic language \citep{pavlopoulos2021semeval} etc. 
Such study, however, has eluded the realm of claims, owning to the lack of quality annotated datasets. This heralds a specialized corpus creation on claim span identification.

To this end, we propose \textbf{\dataset} (\textbf{C}laim \textbf{U}nit \textbf{R}ecognition in \textbf{T}weets), a large-scale, claim span annotated Twitter corpus. We also present several baseline models for solving claim span identification as a token classification task and evaluate them on \dataset. Furthermore, we introduce \textit{claim descriptions}, which are generic prompts aimed to assist the model in focusing on the most significant regions of the input text using explicit instructions on what to designate as a `\textit{claim}'. They are elucidated later in detail.
Finally, we benchmark our dataset with \textbf{\model} (\textbf{D}escription \textbf{A}ware \textbf{R}oBERTa), a plug-and-play adapter-based variant of RoBERTa \citep{liu2019roberta}, endeavored to infuse the Pre-trained Language Model (PLM) with the description information. Empirical results attest that \model\ outperforms the conventional baselines and generic PLMs for our task consistently across various metrics.

\paragraph{\bf Contributions.} Through this work, we make the following tangible contributions: 

\vspace{-0.1cm}
\begin{enumerate}[leftmargin=*]
\setlength{\itemsep}{0 pt}
    \item \textbf{Formulation of a novel problem statement}: We propose the novel task of \textit{Claim Span Identification} that aims to identify argument units of claims in the given text. 
    \item \textbf{Claim span identification dataset and extensive annotation guidelines}: We posit a large-scale Twitter dataset, the first of its kind, with $7.5k$ claim span annotated tweets, to placate the absence of the annotated dataset for claim span identification. Additionally, we develop comprehensive annotation guidelines for the same. 
    \item \textbf{Claim span identification system}: We propose a robust claim span identification framework based on \textit{Compositional De-Attention} (\textit{CoDA}) and \textit{Interactive Gating Mechanism} (\textit{IGM}).
    \item \textbf{Extensive evaluation and analysis}: We evaluate our model against different baselines to confirm sizable improvements over them. We also report thorough qualitative and quantitative analysis along with the ablation studies. 
\end{enumerate}

\paragraph{\bf Reproducibility.} We release our dataset (\dataset) and source code for \model\ publicly at \url{https://github.com/LCS2-IIITD/DABERTA-EMNLP-2022}. 

\section{Why Claim Span Identification?}
\label{sec:motivation}
As stated in Section \ref{sec:intro}, we hypothesize that claim span identification would aid fact-checkers to quickly segregate claim-ridden content from the rest of the post. Moreover, we suppose that it will be a propitious precursor for claim verification and fact-checking, facilitating better retrieval of relevant evidences. We back our hypothesis with a small experiment of evidence-based document retrieval. We collect 50 random samples from \dataset, along with their corresponding ground-truth claim spans. Further, for both the tweets and the claim spans, we extract top-$k$ relevant articles from a knowledge-base leveraging the traditional retrieval system, BM25 \cite{robertson1995okapi}. We use the recently released publicly available CORD19 corpus \cite{wang-etal-2020-cord} to retrieve factual documents. Finally, we present retrieved documents to three evaluators and ask them to mark whether or not the retrieved shreds of evidence are relevant to the given input tweet/span from our dataset. All three annotators label each text-evidence pair independently. Eventually, to obtain the final relevancy score, majority voting is employed. We obtain a high inter-annotator score (\textit{Fleiss Kappa}) of 0.63 and 0.67 for tweets and spans, respectively.

\begin{table}[h]
\centering
\scalebox{0.74}{
\begin{tabular}{l|c|c|c|c} 
\cline{1-5}
\textbf{Input} & \textbf{P@5} & \textbf{P@3} & \textbf{\textit{n}DCG@5} & \textbf{\textit{n}DCG@3} \\ 
\hline
\textbf{Tweets} & 0.3922      & 0.2745 &  0.2733 & 0.2280 \\
\textbf{Spans}  & \textbf{0.4407}      & \textbf{0.3390} & \textbf{0.3038} & \textbf{0.2521}\\ 
\hline
\end{tabular}}
\vspace{-0.5em}
\caption{nDCG@$k$ and P@$k$ scores for tweet and spans using BM25 retrieval system and CORD19 dataset.} 
\label{tab:bm25-scores}
\vspace{-1em}
\end{table}

We compare the performance of tweet-based and span-based retrievals in terms of precision (P) and normalized Discounted Cumulative Gain (nDCG) scores and report them in Table~\ref{tab:bm25-scores}. For comparison, we consider two different top-$k$ settings ($k$=3 and 5). We begin by examining the retrieval performance using P@$k$, which measures the fraction of relevant documents extracted in the top-$k$ set. Span-based document retrieval consistently improves precision scores when compared to tweets. For nDCG@5, we discover that span-based retrieval outperforms tweet-based retrieval by more than $3\%$. When we limit the retrieval depth to 3, we see a similar pattern. 
This, in turn, demonstrates that entire posts contain much extraneous information, frequently impeding the performance of evidence retrieval systems that are a prerequisite for both automated and manual fact-checking. In summary, we reinforce that our hypothesis positively stands true, as span-based document retrieval results in a better score for precision as well as nDCG. This attests to the task's feasibility and importance in the realm of claims.

\section{Related Work}

\paragraph{\textbf{Claims on Social Media.}}
The prevailing research on claims could be cleft into three categories -- claim detection \cite{levy2014context, chakrabarty2019imho, gupta-etal-2021-lesa}, claim check-worthiness \cite{jaradat2018claimrank, wright2020claim}, and claim verification \cite{zhi2017claimverif, hanselowski2018ukp, soleimani2020bert}. \citet{bender-etal-2011-annotating} pioneered the efforts in claim detection by introducing the \textit{AAWD} corpus. Subsequent studies largely relied on using linguistically motivated features such as sentiment, syntax, context-free grammars, and parse-trees  \cite{rosenthal2012detecting, levy2014context, lippi2015context}. 

Recent works in claim detection have engendered the use of large language models (LMs). \citet{chakrabarty2019imho} re-enforced the power of fine-tuning, as their ULMFiT LM, fine-tuned on a large Reddit corpus of about 5M opinionated claims, showed notable improvements in claim detection benchmark. \citet{gupta-etal-2021-lesa} proposed a generalized claim detection model for detecting claims independent of its source. They handled structured and unstructured data in conjunction by training a blend of linguistic encoders (POS and dependency trees) and a contextual encoder (BERT) to exploit the input text's semantics and syntax. As LMs account for significant computational overheads, \citet{sundriyal2021desyr} addressed this quandary and proposed a lighter framework that attempted to fabricate discernible feature spaces. The  \textit{CheckThat! Lab's} \textit{CLEF-$2020$} shared task \cite{barron2020checkthat} has garnered the attention of several researchers. \citet{williams2020accenture} won the task by fine-tuning the RoBERTa \cite{liu2019roberta} accentuated by mean pooling and dropout. \citet{nikolov2020team} ranked second with their out-of-the-box RoBERTa vectors supplemented with Twitter meta-data.

\paragraph{\textbf{Span Identification.}}
\citet{zaidan2007using} introduced the concept of rationales, which highlighted text segments that supported their label's judgment. \citet{trautmann2020fine} released \textit{AURC-$8$} dataset with token-level span annotations for the argumentative components of stance along with their corresponding label. \citet{mathew2020hatexplain} proposed a quality corpus for explainable hate identification with token-level annotations. The \textit{SemEval} community has initiated fine-grained span identification concerning other domains of argument mining such as toxic comments \citep{pavlopoulos2021semeval} and propaganda techniques \citep{da-san-martino-etal-2020-semeval}. These shared tasks amassed many solutions constituting transformers \citep{chhablani2021nlrg}, convolutional neural networks \citep{coope2020span}, data augmentation techniques \cite{rusert2021nlp_uiowa, plucinski2021ghost}, and ensemble frameworks \citep{zhu2021hitsz, nguyen2021s}. \citet{wuhrl2021claim} resembled the closest study to ours, wherein they compiled a corpus of around $1.2k$ biomedical tweets with claim phrases. 

In summary, existing literature on claims concentrates entirely on sentence-level claim identification and does not investigate on eliciting fine-grained claim spans. In this work, we endeavor to move from coarse-grained claim detection to fine-grained claim span identification. 
We consolidate a large manually annotated Twitter dataset for claim span identification task and benchmark it with various baselines and a dedicated description-based model. 


\section{Dataset}
Over the past few years, several claim detection datasets have been released \cite{rosenthal2012detecting, chakrabarty2019imho}. However, none of these corpora come with claim-based rationales that quantify a post as a claim. To bridge this gap, we propose \textbf{\dataset} ({\textbf{C}laim \textbf{U}nit \textbf{R}ecognition in \textbf{T}weets}), a large scale Twitter corpus with token-level claim span annotations. 

\paragraph{Data Selection.}
\label{sec:data-selection}
We annotate claim detection Twitter dataset released by \citet{gupta-etal-2021-lesa} for our task. However, the guidelines 
they presented have certain reservations, wherein they do not explicitly account for benedictions, proverbs, warnings, advice, predictions, and indirect questions. As a result, tweets such as \textit{`Dear God, Please put an end to the Coronavirus. Amen'} and \textit{`@FLOTUS Melania, do you approve of ingesting bleach and shining a bright light in the rectal area as a quick cure for \#COVID19? \#BeBest'} have been mislabeled claims. This prompted us to extend the existing guidelines and introduce a more exclusive and nuanced set of definitions based on claim span identification. We present details of the extended annotation guidelines and guideline development procedure in  \hyperref[sec:annotations]{\textcolor{blue}{Appendix (\ref{sec:annotations})}}. In total, we annotated $7555$ tweets from the Twitter corpus by \citet{gupta-etal-2021-lesa} which met our guidelines. 

\begin{table}
    \centering
    \scalebox{0.74}{
    \begin{tabular}{l|c|c|c}
    \hline
      \bf Dataset & \bf Train &\bf Test &\bf Validation\\ 

        \hline
      \bf  Total no. of claims &6044 &755 &756 \\
        \bf Avg. length of tweets &27.40 &26.93 &27.29 \\
        \bf Avg. length of spans &10.90 &10.97 &10.71 \\
        \bf No. of span per tweet &1.25 &1.20 &1.27 \\
        \bf No. of single span tweets &4817 &629 &593 \\
        \bf No. of multiple span tweets &1201 &121 &161 \\
        \hline
        
    \end{tabular}
    }
    \vspace{-0.5em}
    \caption{Dataset statistics. All the lengths are in tokens.} 
    \label{tab:dataset-stats2}
    \vspace{-1em}
\end{table}


\paragraph{Dataset Statistics and Analysis.}
\label{sec:dataset-statistics}
We segment \dataset\ into three partitions -- training set, validation set, and test set, in the split of $80$:$10$:$10$. Dataset related statistics are given in Table~\ref{tab:dataset-stats2}. One important point to note here is that while a claim tweet is typically $27$ tokens long, a claim span is only around $10$ tokens long. This implies that the claim-ridden tweets have a lot of extraneous information. 
Arguments can also perhaps comprise several claims that may or may not be related to each other. Around $19\%$ of the claim tweets in our dataset contain multiple claim spans. As a result, in total, we obtain $9458$ claim spans from $7555$ tweets.  
We observe that the majority of the tweets contain single claims. 
Out of $7555$ tweets, $6039$ include a single claim, demonstrating that the majority of tweets contemplate single assertions at a time.

\section{Proposed Methodology}
\begin{figure*}
\includegraphics[width=\textwidth]{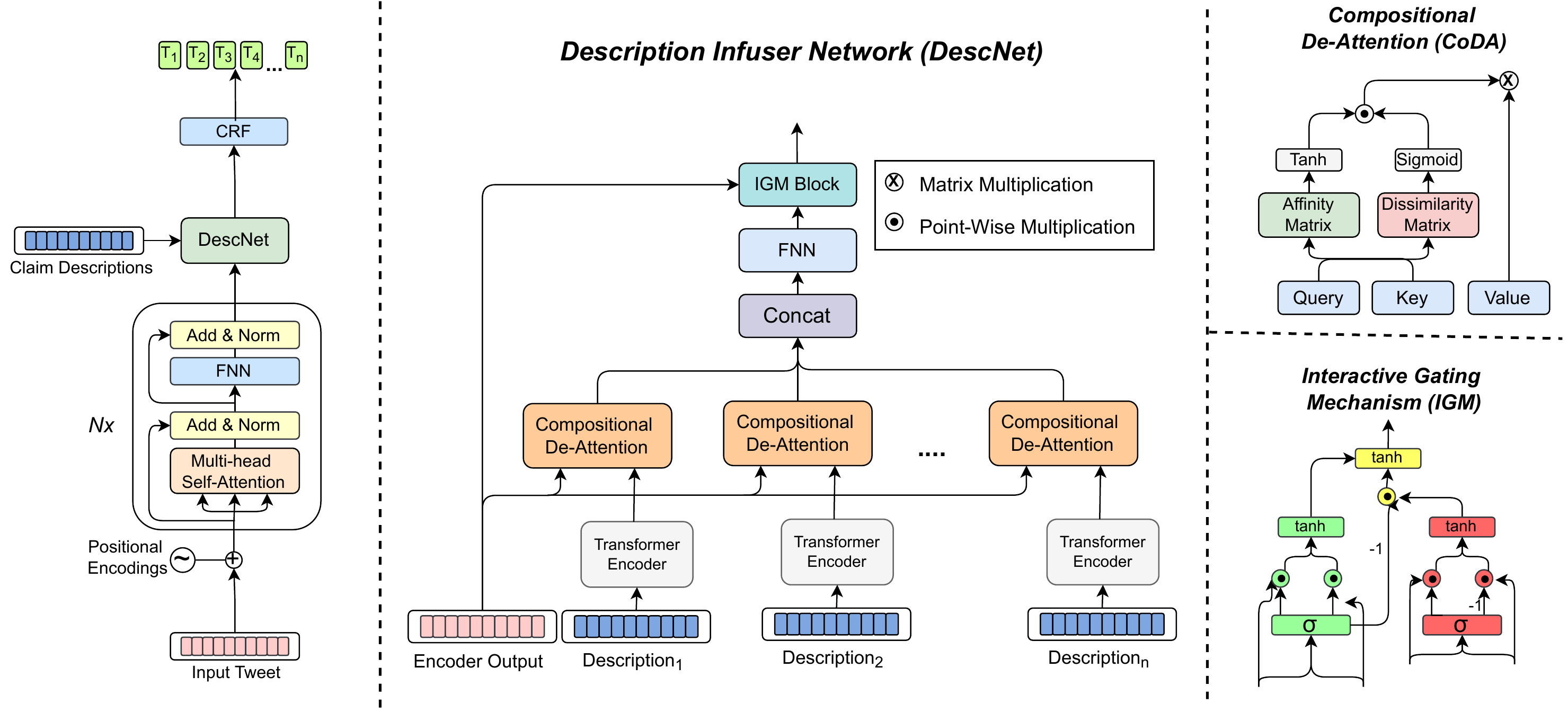}
\vspace{-1.5em}
\caption{A schematic diagram of \model\ for the claim span identification. $\odot$ represents point-wise multiplication, and $\otimes$ represents matrix multiplication. }
\label{fig:model}
\end{figure*}

\label{sec:model}
In this section, we outline \model\ and its intricacies. The main aim is to seamlessly coalesce critical domain-specific information into Pre-trained Language Models (PLM). 
To this end, we introduce \textit{Description Infuser Network} (\textit{DescNet}), a plug-and-play adapter module that conditions the LM representations with respect to the handcrafted descriptions. The underlying principle behind this theoretical formalization is to link a claim span to a claim description to guide the model on what to focus on explicitly. As shown in Figure \ref{fig:model}, DescNet houses two sub-components, namely, \textit{Compositional De-Attention block} (\textit{CoDA}) and \textit{Interactive Gating Mechanism} (\textit{IGM}). 
The particulars of each component are delineated in the following sections.

\begin{table*}[ht]
    \centering
    \scalebox{0.75}{
    \begin{tabular}{p{21em}|p{31em}}
    \hline
       \bf Claim Description & \bf Example\\ 

        \hline
        Texts in the tweet mentioning statistics, dates or numbers & Another case for more testing for \#coronavirus! \textit{Blood tests show 14\% of people are now immune to covid-19 in one town in Germany} https://t.co/MVOq3nc4hn \\ 
        \hline 
        Texts in the tweet that negate a possibly false claim & No! \textit{\#Bleach won't cure \#COVID19. Disinfectants can't kill the \#coronavirus in your body}. In fact, they will hurt you. If you or someone you know has been exposed to bleach, call Poison Control for help (1-800-222-1***). https://t.co/DtIfi77vLz https://t.co/9MxSFoVM0L \\ 
        \hline
        Texts in the tweet made in sarcasm or humour & @username I think the \textit{cure to coronavirus is a 6 pack of corona} only. yeah\\ 
        \hline
        Texts in the tweet containing opinions that have societal implications & @username @username I think \textit{it’s a bio weapon made by China} so I’m not surprised it has a lot of carriers. \\ 
        \hline
        Texts in the tweet in the form of conditional statement & \textit{if you smoke weed you are immune to coronavirus}\\ 
        \hline
        Texts in the tweet containing a quote from someone & The president said \textit{injecting disinfectant into the body can cure the virus}. What in the holy hell? And @Lysol issued a statement that people should not ingest Lysol. WTF? \#Covid\_19 \#lysol \#DontDrinkLysol\\
        \hline

    \end{tabular}
    }
    \caption{Examples of handcrafted claim descriptions, along with some aligning examples. Claim spans are highlighted in italics.} 
    \label{tab:descriptions}
    \vspace{-5mm}
\end{table*}

\paragraph{Claim Descriptions.}
\label{sec:descriptions}
Before delving into \textit{CoDA} and \textit{IGM}, we first examine Claim Descriptions, which are the cornerstone of the proposed model. Claim Descriptions are handcrafted templates that guide the model where to concentrate its focus. The inclusion of claim description encourages the model to focus on the most essential phrases in the input tweet, which may be thought of as guided attention that leads to increased performance. We judiciously curated our claim descriptions in accordance with the annotation guidelines for claims and non-claims offered by \citet{gupta-etal-2021-lesa}. In Table \ref{tab:descriptions} we list some of the claim descriptions along with the claims that they most align with. It is noteworthy that a claim can align with more than one claim descriptions as well.

\paragraph{Overview of PLMs for Token Classification.}
To begin with the details of the proposed framework, \model, we present the working of PLMs for the token classification task. PLMs such as BERT \citep{devlin-etal-2019-bert}, DistilBERT \citep{sanh2019distilbert}, and RoBERTa \citep{liu2019roberta} are widely used for various downstream NLP tasks owning to their strong contextual language representation capabilities and fine-tuning ease. As the input to these PLMs, each $i^{th}$ input text is first tokenized into a sequence of sub-word embeddings $ X_i \in \mathbb{R}^{N \times d} $, where $N$ is the maximum sequence length and $d$ is the feature dimension. Then a positional embedding vector $ PE_{pos} \in \mathbb{R}^{N \times d}$ is added to the token embeddings in a pointwise fashion 
to retain the positional information \citep{vaswani2017attention}.

The vector $ Z_i \in \mathbb{R}^{N \times d}$, hence obtained, is fed to a stack of transformer encoder blocks. Each encoder block is a modular unit consisting of two sub-layers: $(i)$ \textit{Multi-Headed Self-Attention}, and $(ii)$ \textit{Feed-Forward Network}. Furthermore, each sub-layer contains a residual connection, followed by dropout and layer normalization. For the task of token classification, the output of the last encoder layer is passed to a CRF layer \citep{lafferty2001conditional}. This modularity of PLMs enables easy integration of adapter modules in their architecture for making these PLMs task-specific and domain-dependent. We choose RoBERTa \cite{liu2019roberta} as our backbone network as it is the best-performing baseline (see Table ~\ref{tab:results}). 

\paragraph{Description Infuser Network (\textit{DescNet}).}
DescNet is designed to facilitate deep semantic interaction among the input text and claim descriptions, and help underline the key fragments of claims. 
It consists of precisely engineered components of CoDA and IGM, each devised to augment the process of claim span identification.

To put formally, consider $D = \{d_1, d_2, ..., d_m\}$ as the set of $m$ claim descriptions and $T = \{t_1, t_2, ..., t_n\}$ as the corpus of $n$ input texts. The description representations are extracted from pre-trained RoBERTa  \citep{liu2019roberta} and passed through a transformer encoder layer. To begin with, each $i^{th}$ PLM generated vector $Z_i \in \mathbb{R}^{N \times d}$ of input text $t_i$ interacts with each $j^{th}$ description vector $D_j \in \mathbb{R}^{M \times d}$ via the CoDA block. Here the vector $Z_i$ forms the query, which is processed against the vector $D_j$ acting as the key and value (Equation~\ref{eq:coda-1}).

\vspace{-0.5em}
\begin{small}
\begin{equation}
    \label{eq:coda-1}
    Z_{ij}^{C} = CoDA(Z_i, D_j)D_j
\end{equation}
\end{small}

All such compositionally manipulated vectors $Z_{ij}^C$, after interacting with each $j^{th}$ description vectors are concatenated and passed through a dropout layer before going through a non-linear transformation for dimensionality reduction (Equation~\ref{eq:coda-after-2}). The resultant vector $Z_i'$ along with the vector $Z_i$ is passed to the \textit{IGM} module to extract the semantically appropriate features pertinent for fine-grained claim span identification (Equation~\ref{eq:gain}).

\vspace{-0.5em}
\begin{small}
\begin{eqnarray}
\label{eq:coda-after-2}
    Z_i' &=& Concat(Z_{i1}^{C}, ... ,Z_{im}^{C}) \\
  \hat{Z_i} & = & IGM(Z_i'W, Z_i) \label{eq:gain}
\end{eqnarray}
\end{small}
The vector $\hat{Z_i}$ is 
then passed to a CRF layer. 

\paragraph{Compositional De-Attention Block (\textit{CoDA}).}

The traditional narrative on attention mechanism \citep{BahdanauCB14, parikh-etal-2016-decomposable, seo2016bidirectional, vaswani2017attention} is heavily biased on the use of \textit{Softmax} operator where the attention weights are always bounded between $[0, 1]$. Such a convex weighted addition scheme allows the vectors to only contribute in an additive manner. To counter this bottleneck, \citet{tay2019compositional} devised a quasi-attention technique that enables learning of additive as well as subtractive attention weights, allowing the input vectors to add to $(+1)$, not contribute to $(0)$, and even subtract from $(-1)$ the output vector. They decomposed the original Softmax-based self-attention as pointwise multiplication between two matrices as shown in Equation~\ref{eq:coda}, where $G(.)$ is the negative pointwise $L_1$ distance between query $Q$ and key $K$.  

\vspace{-0.5em}
\begin{small}
\begin{eqnarray}
\label{eq:coda}
    A_{quasi} = \Bigg(tanh(\frac{QK^T}{\sqrt{d_k}})\odot \sigma(\frac{G(Q, K)}{\sqrt{d_k}})\Bigg)V  
\end{eqnarray}
\end{small}

We adopt this quasi-attention strategy to promote more meaningful interaction between the input text and claim descriptions and generate more precise claim-relevant representations. 

\paragraph{Interactive Gating Mechanism (\textit{IGM}).}
To further distinguish salient tokens inclusive in claim spans, we posit \textit{Interactive Gating Mechanism}. To begin with, the vectors $Z_i$ and $Z_i'$ are max pooled to obtain $Z_{ip}$, $Z_{ip}'$ $\in \mathbb{R}^{d}$. These vectors are passed through a series of gates, the first of them being the \textit{conflict gate} $C$, aimed at capturing the semantically conflicting features in $Z_i$ and $Z_i'$ (Equation~\ref{eq:conflict}).

\vspace{-0.5em}
\begin{small}
\begin{align}
     \mu_c &= \sigma(Z_{ip}W_{c1} + Z_{ip}'W_{c2}+ b_{c1})\\
     \label{eq:conflict}
     C &= \tanh(Z_{ip}\odot\mu_c W_{c3}+ Z_{ip}'\odot(1-\mu_c)W_{c4}+b_{c2})
\end{align}
\end{small}



The \textit{refine gate} $R$, on the other hand, endeavors to capture the semantically similar features between $Z_{ip}$ and $Z_{ip}'$ (Equation~\ref{eq:refine}).

\vspace{-0.5em}
\begin{small}
\begin{eqnarray}
    \label{eq:mur}
     \mu_r &=&  \sigma(Z_{ip}W_{r1} + Z_{ip}'W_{r2}+ b_{r1}) \\
     \label{eq:refine}
     R &=& \tanh(Z_{ip}\odot\mu_r W_{r3} + Z_{ip}' \odot\mu_rW_{r4}+ b_{r2})
\end{eqnarray}
\end{small}

To congregate the conflicting and similar semantic representations spawned by the gates $C$ and $R$, we employ an adaptive gating scheme to retain maximum differential information from each gate. It is given by Equation \ref{eq:adapt}. 

\vspace{-0.5em}
\begin{small}
\begin{eqnarray}
    A &=& R + (1-\mu_r)\odot C\\
    \label{eq:adapt}
    \hat{Z}_i &=& tanh(AW_a + b_a) \odot Z_i 
\end{eqnarray}
\end{small}

Finally, this vector $\hat{Z_i}$ is 
passed to a CRF layer for token classification.

\begin{table*}
\centering
\resizebox{0.99\textwidth}{!}{
\begin{tabular}{l|c|c|c|c|c|c|c|c|c|c|c|c|c} 
\hline
\multirow{2}{*}{\bf Model Name} & \multirow{2}{*}{\bf F1} & \multirow{2}{*}{\bf P} & \multirow{2}{*}{\bf R} & \multicolumn{3}{c|}{\bf F1}  & \multicolumn{3}{c|}{\bf Precision} & \multicolumn{3}{c|}{\bf Recall} & \multirow{2}{*}{\bf DSC}        \\ \cline{5-13}
&                           &                            &                         & \multicolumn{1}{c|}{\bf B} & \multicolumn{1}{c|}{\bf I} & \multicolumn{1}{c|}{\bf O} & \multicolumn{1}{c|}{\bf B} & \multicolumn{1}{c|}{\bf I} & \multicolumn{1}{c|}{\bf O} & \multicolumn{1}{c|}{\bf B} & \multicolumn{1}{c|}{\bf I} & \multicolumn{1}{c|}{\bf 0} &  \\ 
\hline
\bf CNN+CRF                                                      & 0.6635                    & 0.6709                     & 0.6947                  & 0.3877                 & 0.6766                 & 0.9263                 & 0.3952                 & 0.6725                 & \textbf{0.9450}         & 0.3953                 & 0.7718                 & 0.9171              &   0.6964 \\

\bf BiLSTM+CRF                                                       & 0.6825                    & 0.6928                     & 0.7048                  & 0.4401                 & 0.6717                 & \textbf{0.9356}        & 0.4653                 & 0.6703                 & 0.9428                 & 0.4302                 & 0.7459                 & \textbf{0.9382}             & 0.6884    \\
\hline

\bf DistilBERT              & 0.7645   & 0.7811    & 0.8068 & 0.6677 & 0.8164 & 0.7510  & 0.6560        & 0.7979       & 0.8310        & 0.7227    & 0.8989    & 0.7402     & 0.8277 \\

\bf BERT                    & 0.7807   & 0.7996    & 0.8154 & 0.6900   & 0.8266 & 0.7699 & 0.6863       & 0.8163       & 0.8403       & 0.7302    & 0.8971    & 0.7634     & 0.8356\\

\bf SpanBERT                & 0.7914   & 0.8093    & 0.8182 & 0.6971 & 0.8299 & 0.7901 & 0.7047       & 0.8384       & 0.8271       & 0.7203    & 0.8724    & 0.8048   & 0.8377 \\

\bf RoBERTa                 & 0.8020    & 0.8163    & 0.8337 & 0.7221 & 0.8297 & 0.7942 & 0.7165       & 0.8371       & 0.8351       & 0.7624    & 0.8764    & 0.8022     & 0.8399 \\

\hline

\bf DistilBERT + CRF                                     & 0.8288                    & 0.8581                     & 0.8526                  & 0.8722                 & 0.8148                 & 0.7431                 & 0.8914                 & 0.7852                 & 0.8400                 & 0.8621                 & 0.9181                 & 0.7219             & 0.8222    \\

\bf BERT + CRF                                            & 0.8368                    & 0.8631                     & 0.8556                  & 0.8531                 & 0.8284                 & 0.7666                 & 0.8781                 & 0.8101                 & 0.8375                 & 0.8408                 & 0.9042                 & 0.7597                & 0.8343  \\

\bf SpanBERT + CRF   & 0.8390    & 0.8625    & 0.8562 & 0.8507 & 0.8253 & 0.7806 & 0.8742       & 0.8221       & 0.8302       & 0.8394    & 0.8827    & 0.7867   & 0.8316  \\

\bf RoBERTa + CRF                                       & 0.8457   & 0.8706    & 0.8635 & 0.8613 & 0.8340  & 0.7805 & 0.8874       & 0.8301       & 0.8321       & 0.8485    & 0.8972    & 0.7841 & 0.8402     \\

\hline

\bf NLRG & 0.7494 &	0.7750 &	0.7832 &	0.6584 &	0.7892 &	0.7398 &	0.6631 &	0.7891 &	0.8119 &	0.6805 &	0.8600	& 0.7486  & 0.8087\\

\bf HITSZ-HLT   &   0.7758	& 0.7966 & 0.8037 &	0.6754 &	0.8201 &	0.7780	& 0.6834 &	0.8230	& 0.8291 &	0.6978 &	0.8743 &	0.7850  & 0.8314\\

\hline

\rowcolor[rgb]{0.839,0.882,1}\textbf{\model} & \textbf{0.8604}           & \textbf{0.8814}            & \textbf{0.8789}         & 0.9035        & \textbf{0.8354}        & 0.7816                 & 0.9205        & \textbf{0.8242}                 & 0.8379                 & 0.8950         & \textbf{0.9044}        & 0.7771          &  \textbf{0.8433}    \\
\hline
\bf  {$\quad-$ \{\textit{IGM}\}} 
& 0.8539   & 0.8795    & 0.8768 & \textbf{0.9121} & 0.8310  & 0.7563 & \textbf{0.9258}       & 0.8017       & 0.8473       & \textbf{0.9051}    & 0.9277    & 0.7358          &  0.8401     \\

\bf  {$\quad-$ \{\textit{CoDA}\} $+$ \{\textit{DPA}\}}           & 0.8558   & 0.8788    & 0.8738 & 0.8886 & 0.8319 & 0.7821 & 0.9086       & 0.8184       & 0.8439       & 0.8785    & 0.9032    & 0.7752 & 0.8416    \\

\hline
\end{tabular}}
\vspace{-0.5em}
\caption{Experimental results of \model, its variants (last two rows), and baselines. DSC, P, and R denote Dice Similarity Coefficient, Precision, and Recall respectively.} 
\vspace{-1em}
\label{tab:results}
\end{table*}

\section{Experiments and Results}
\paragraph{Baseline Models.} 

We employ the following baseline systems. $\triangleright$ \textbf{CNN+CRF}: A Convolutional Neural Network (CNN) trained with GloVe \citep{pennington-etal-2014-glove} and a CRF head on top. $\triangleright$ \textbf{BiLSTM+CRF} \cite{huang2015bidirectional}: A sequence labeling model comprising Bidirectional Long Short-Term Memory (BiLSTM) and CRF layer. $\triangleright$ \textbf{BERT} \cite{devlin-etal-2019-bert}: A bidirectional transformer-inspired auto-encoder language model fine-tune for our span identification task. $\triangleright$ \textbf{DistilBERT} \cite{sanh2019distilbert}: A smaller, faster, and lighter version of BERT fine-tune on our dataset for the task at hand. $\triangleright$ \textbf{SpanBERT} \cite{joshi2020spanbert}: An enhanced version of the BERT trained on span prediction objective. $\triangleright$ \textbf{RoBERTa} \cite{liu2019roberta}: A robustly optimized BERT approach, RoBERTa, is a variant of BERT with improved training methodology. We fine-tune it on our dataset. $\triangleright$ \textbf{NLRG} \cite{chhablani2021nlrg}: A system proposed at SemEval-2021 Task 5 on toxic span detection \citep{pavlopoulos2021semeval}. It is a combination of SpanBERT and RoBERTa where the former model is used for predicting the span start and end, while the latter is used for token classification. $\triangleright$ \textbf{HITSZ-HLT} \cite{zhu-etal-2021-hitsz}: The system topped the SemEval-2021 task on toxic span detection. They approached the task as a combination of sequence labeling and span extraction and proposed an ensemble of three BERT-based models.

\paragraph{Evaluation Metrics.}
In concordance with \citet{pavlopoulos2021semeval}, we evaluate the performance of all the systems, based on token-level precision (P), recall (R), and F1 scores.
To further put a lens over how the models fare for different token types, we calculate the micro-level precision, recall, and F1 score for each of the \textit{`B'}, \textit{`I'}, and \textit{`O'} tokens.\footnote{Each token in the tweet is \textit{BIO} \textit{(Begin-Inside-Outside)} encoded to mark the claim spans \citep{ramshaw1999text}.} 
Lastly, to quantify the number of tokens included in the spans, we also report the Dice Similarity Coefficient (DSC) \cite{dice1945measures}. 

\paragraph{Performance Comparison.} We summarize our collated results 
in Table \ref{tab:results}. Evidently, \model\ outperforms all the baseline systems against majority of the evaluation metrics.\footnote{Data preprocessing and best hyperparameter configurations are presented in \hyperref[sec:hyperparameter]{\textcolor{blue}{Appendix (\ref{sec:hyperparameter})}}.} We analyze all the systems based on the following questions.

\noindent \underline{\textit{How accurately do the models predict?}}
To gauge how well each model performs for the token classification task, we monitor precision, recall, and F1 scores. As it can be inferred from Table~\ref{tab:results}, the traditional word embedding-based deep learning models of CNN and BiLSTM give the poorest token classification performance. An appreciable improvement of about $10$-$14\%$ across all three metrics is observed when we move from the classical deep learning architectures to the transformer-based models of DistilBERT, BERT, SpanBERT, and RoBERTa. This underlines the importance of using contextual word embeddings and transformer-based architectures for the task at hand. The addition of the CRF layer further amplifies the performance of these models. 
SpanBERT also fares better than BERT as it is trained using span prediction objective. We also notice that employing the CRF layer results in a somewhat better balance of precision and recall when compared to using a basic linear layer. 
The ensemble-based models of NLRG and HITSZ-HLT also give admissible results for our task. 
Our proposed model, \model, surpasses all the models in terms of the 
precision, recall, and F1 scores. An improvement of about $1.5\%$ is observed between RoBERTa and \model\ in terms of these metrics. This justifies the inclusion of \textit{claim descriptions} that amalgamate domain-specific semantic information into RoBERTa architecture via the deftly crafted adapter module. In summary, we see that all the models show a good trade-off between precision and recall.

\noindent \underline{\textit{Are the models aggressive or defensive?}}
Observing the precision, recall, and F1 scores for each of the \textit{`B'}, \textit{`I'}, and \textit{`O'} tags, as shown in Table \ref{tab:results}, we get an idea of how aggressive or defensive the models are at predicting claim spans. CNN and BiLSTM show considerable resistance in predicting the claim spans, as evidenced by high precision, recall, and F1 scores for the token \textit{`O'} and less for the tokens \textit{`B'} and \textit{`I'}. The BERT-based models show a sizable improvement of about $22\%$ and $15\%$ for predicting tokens \textit{`B'} and \textit{`I'}, respectively, over the traditional deep learning models. The addition of CRF layers further bolsters the predictive power for the token \textit{`B'}. \model\ offers an improvement of about $4$-$5\%$ over its traditional counterpart for predicting the token \textit{`B'}. Upon close inspection, we observe that the ranges of precision, recall, and F1 scores for predicting the tokens \textit{`I'} and \textit{`O'} vary by not more than $3\%$. However, the predictive power for the token \textit{`B'} varies vastly by about $25\%$. Hence, we hypothesize that the inclusion of \textit{descriptions} makes our model cognizant of the syntactic and semantic constructs of claims.

\begin{figure}
    \includegraphics[width=0.46\textwidth]{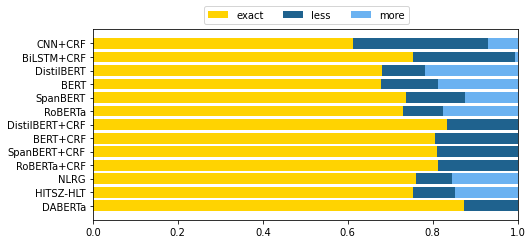}
    \vspace{-0.5em}
    \caption{A comparative study among \model\ and baselines. The horizontal bar signifies the  ration of number of predicted spans and number of gold spans.}
    \vspace{-1.5em}
    \label{fig:span-counts}
\end{figure}

\noindent \underline{\textit{How the models behave for multiple spans?}}
Figure~\ref{fig:span-counts} illustrates how well the models identify multiple spans. It is observed that the models of CNN and BiLSTM find it challenging to identify multiple spans. The transformer-based models with a linear head tend to predict more claim spans in the tweet than required. This issue is mitigated when the linear head is replaced with a CRF layer. Still, these models can identify roughly only $80\%$ of the time the occurrence of multiple spans. On the other hand, our model, \model, correctly predicts multiple spans almost more than $85\%$ of the time. Moreover, it does not predict more claim spans than required. Thus, the addition of domain-specific \textit{claim descriptions} appropriately guides \model\ in identifying the correct occurrence of spans.

\paragraph{Ablation Study.}
Table~\ref{tab:results} also reports the ablation studies. Replacing \textit{CoDA} with a naïve \textit{Dot-Product Attention} (\textit{DPA}), we observe a drop in the performance across almost all the metrics. Amongst all, the performance drop in predicting the token \textit{`B'} is the most prominent ($\sim1.5\%$ across precision, recall, and F1). Thus, we conjecture that the quasi-attention mechanism is better able to spot the starting of a claim fragment than \textit{DPA}. When \textit{IGM} is removed, the performance for predicting token \textit{`B'} slightly improves. However, it leads to a decrease in the predictive power for \textit{`O'} token  ($\sim2.5\%$ in F1). Therefore, the combination of \textit{CoDA} and \textit{IGM} obtains the most balanced performance.

\paragraph{Hyper-parameter Tuning.}
\label{sec:hyperparameter}
We utilize the \textit{base} version of RoBERTa \citep{liu2019roberta} to propose \model. The model is trained end-to-end using the Adam optimizer \citep{kingma2014adam}, learning rate of $4e-5$, and batch size of 32 for 20 epochs with early stopping if the dice score does not improve after 5 epochs. We used the Nvidia Tesla v100 32 GB GPU. The hyper-parameter tuning is done with respect to the validation dataset.


\begin{figure}[h]
    \centering
    \includegraphics[width=0.45\textwidth]{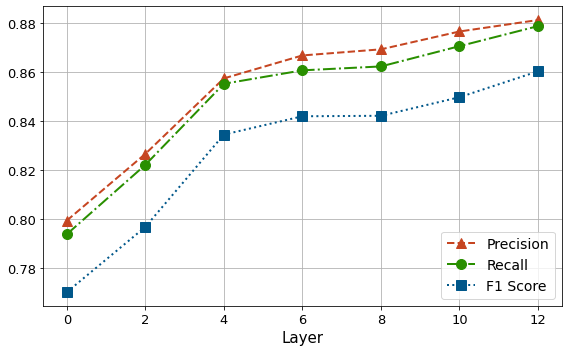}
    \vspace{-0.5em}
    \caption{Performance of \model\ when the adapter module is inserted at different layers of RoBERTa.}
    \label{fig:layerwise-ablation}
    \vspace{-1em}
\end{figure}

Figure \ref{fig:layerwise-ablation} reflects the effect of integrating the adapter \textit{DescNet} at different layer of RoBERTa. It is observed that the performance consistently increases as the integration is done at a higher level of RoBERTa layers. This is admissible as studies on probing the PLM layers suggest that different layers encode distinct linguistic properties \citep{tenney-etal-2019-bert}. Furthermore, evidence by \citet{peters-etal-2018-dissecting} suggests that the lower layers of a language model encode the syntactic information, whereas the higher layers capture the complex semantics. As we strive to employ deep semantic interaction between the PLM representations and the \textit{claim descriptions}, our results are consistent with their findings.

\paragraph{Error Analysis.}
\begin{table*}[!ht]
\centering
\small
\resizebox{\textwidth}{!}{
\begin{tabular}{l|c|p{50em}} 
\hline
& \textbf{Model} & \textbf{Tweet}  \\ 
\hline
& \textbf{\textit{Gold}} & Gold Truly sobering analysis: \hlcorrect{\textbf{US more vulnerable than many countries to \#coronavirus}} owing to combination of high numbers of uninsured, many w/o paid sick leave, and a leadership that has downplayed the challenge while not preparing the country for it.      
\\ 
\cdashline{2-3}

1  & \textbf{\textit{RoBERTa}} & Gold Truly sobering analysis: \hlcorrect{\textbf{US more vulnerable than many countries to \#coronavirus}} owing to combination of \hlwrong{\textit{high numbers of uninsured}}, many w/o \hlwrong{\textit{paid sick leave}}, and a leadership that has downplayed the challenge while not preparing the country for it. 
\\ 
\cdashline{2-3}
& \textbf{\model} & Gold Truly sobering analysis: \hlcorrect{\textbf{US more vulnerable than many countries to \#coronavirus}} owing to combination of high numbers of uninsured, many w/o paid sick leave, and a leadership that has downplayed the challenge while not preparing the country for it.          \\ 
\hline

& \textbf{\textit{Gold}} & Whether made on purpose or not  \hlcorrect{\textbf{\#coronavirus was used by the \#CCP as a bio weapon}}, not only to kill people but to encourage racism among their citizens against foreigners. Especially black people,  \hlcorrect{\textbf{CCP is kicking out black people from hotels even if they dont have covid.}}                         
\\ 
\cdashline{2-3}
2 & \textbf{\textit{RoBERTa}} & Whether made on purpose or not  \hlcorrect{\textbf{\#coronavirus was used by the \#CCP as a bio weapon}},  \hlwrong{\textit{not only to kill people but to encourage racism among their citizens against foreigners}}. Especially black people, CCP is kicking out black people from hotels even if they dont have covid.    
\\ 
\cdashline{2-3}
& \textbf{\model} & Whether made on purpose or not  \hlcorrect{\textbf{\#coronavirus was used by the \#CCP as a bio weapon}}, \hlwrong{\textit{not only to kill people but to encourage racism among their citizens against foreigners}}. Especially black people, \hlcorrect{\textbf{CCP is kicking out black people from hotels even if they dont have covid.}}                       
\\ 
\hline
 & \textbf{\textit{Gold}} & RT @HealtheNews: Can honey, ginger, garlic or turmeric or any other home remedies cure \#Covid19? No, here’s why. 
\\ 
\cdashline{2-3}
3 & \textbf{\textit{RoBERTa}} & RT @HealtheNews: Can \hlwrong{\textit{honey, ginger, garlic or turmeric or any other home remedies cure \#Covid19?}} No, here’s why.  
\\ 
\cdashline{2-3}
& \textbf{\model} & RT @HealtheNews: Can \hlwrong{\textit{honey, ginger, garlic or turmeric or any other home remedies cure \#Covid19?}} No, here’s why. \\ 
\hline
\end{tabular}}
\caption{Error analysis of the outputs. Bold text (green) highlights the correct claim span whereas text in italics (red) represents the mistakes committed by our model, \model, and vanilla RoBERTa as baseline.}
\vspace{-2em}
\label{tab:error-analysis}
\end{table*}

In this section, we manually analyze 
the errors the models are prone to make. Table \ref{tab:error-analysis} highlights randomly sampled tweets from our dataset, \dataset, along with their gold spans and predictions from \model. In addition, we also consider the predictions from the best-performing baseline, RoBERTa, for a fair comparison. 
We analyze the errors committed by both the systems and divide them into three different categories: (i) tweets with a single-claim span, (ii) tweets with claim-like premises, and (iii) tweets with claims that can be inferred from the underlying undertone of the tweet but no explicit span can be marked to highlight the claim-specific connotation, e.g., figurative sentences, satire, indirect questions etc. (Note: For simplicity, we refer to such claims as \textit{implicit} claims.) In the most straightforward situation where the tweets only contain a single claim   \model\ makes more precise predictions than the baseline system as shown in the first example of Table \ref{tab:error-analysis}. We observe that both the models identify the claim-span correctly; however, RoBERTa identifies some unnecessary spans, which trespasses our objective of equipping the fact-checkers with only relevant information.

The second type of error related to spans is the presence of claim-like premises. Claims and premises\footnote{(sub)sentences used to support the concluding claim} are closely related components of argument mining, and differentiating them is strenuous, even for humans. Example $2$ in Table \ref{tab:error-analysis}, exhibits a post containing claim-premise pair. There are two conclusive claims in the tweet -- \textit{ `\#coronavirus was used by the \#CCP as a bio weapon'} and \textit{`CCP is kicking out black people from hotels even if they don't have covid'}. Even though \textit{`not only to kill people but to encourage racism among their citizens against foreigners'} appears to be a claim at first glance, it serves as the premise to support the conclusive part of the arguments brought forward in the tweet. In most cases, we discern that both the systems identify the claim-spans correctly, but they are easily fooled by the premises, hence leaving room for significant improvement in this regard. 

Another prominent class of errors is implicit claims. Extracting the claim-spans in implicit claims is arduous. We observe that both the systems strive to understand the linguistic structure of the implicit claims. For instance, in sample $3$, the user intends to assert that honey, ginger, garlic, or turmeric do not cure COVID19; however, \model\ fails to understand the user's intention and yields the wrong span. We perceive similar behavior from the best-performing baseline, RoBERTa, as well. A plausible reason is the skewed nature of the dataset , which is lopsided with a significant bias toward explicit claims. According to our observations, \model\ outperforms the best-performing baseline system significantly ($\sim4\% ;p<0.0004$).\footnote{We also perform significance t-test on F1 scores per input tweet comparing \model\ and RoBERTa, as best-performing baseline.} Hence, furnishing us with empirical shreds of evidence that \model\ can be efficiently used for claim span identification.

\section{Conclusion}
\vspace{-0.5em}
Through this systematic research, we introduced the novel task of \textit{Claim Span Identification}, which is valuable on various fronts. 
We conducted an evidence-based document retrieval experiment, demonstrating that employing claim spans retrieves more relevant evidence than using the entire tweet. Furthermore, as there exists no specialized corpus for claim span identification, we compiled \textbf{\dataset}, a large-scale Twitter corpus consisting of around $7.5k$ tweets annotated with token-level claim spans. We showed convincing results using various token classification baselines on our dataset. Moreover, we benchmarked \dataset\ with \textbf{\model}, an adapter-based variant of RoBERTa, that encapsulates critical domain-specific information into the pre-trained model via \textit{claim descriptions}. Through extensive qualitative, quantitative, and empirical results, we illustrated how \model\ outperforms the other models on different fronts. Lastly, we also developed an extensive set of annotation guidelines and released them for further research.

\section{Limitations}
Though \model\ yields the state-of-the-art performance in claim span identification; there are a few cases where it falls short. Even for humans, recognizing claim spans in figurative or metaphorical sentences is arduous; consequently, our suggested model also struggles with them. As a result, our future study will focus on boosting the claim span identification performance, especially for such sentences. Our analysis also bestowed that the high resemblance between claims and premises confuses the model, making it difficult to distinguish between the two. \model\ shares the said limitation with other baseline systems as well. As a result, this could be another alluring open challenge to work on.

\section*{Acknowledgement}
The authors would like to thank the Ramanujan Fellowship, SERB.

\bibliography{main}
\bibliographystyle{acl_natbib}

\clearpage
\appendix
\section{Appendix}
\label{sec:appendix}
\subsection{Annotations}
\label{sec:annotations}
\subsubsection{Guideline Development}
\label{sec:guideline-development}
While different frameworks and models of argumentation range in intricacy and claim conceptualization, the claim element is colloquially perceived as a principal component of an argument.  Following \citet{stab2017parsing}, we define the claim as `\textit{the argumentative component in which the speaker or writer conveys the central, contentious conclusion of their argument}'. \citet{aharoni-etal-2014-benchmark} proposed a framework in which an argument is often divided into two parts: claim and premise. The premise, which is another crucial component of an argument, encompasses all shreds of evidence obliged to either corroborate or refute the claim. We confine our corpus to claim components only. However, claims and premises are usually indistinguishable and frequently blend together. As a result, distinguishing them can be challenging, especially when authors use claim-like statements as a premise. 

Due to the highly subjective nature of claims, it is imperative to devise structured annotation guidelines to annotate a new dataset for the claim span identification task. Therefore, after rigorous analysis and discussion, we established an initial set of annotation guidelines. To acclimate better with the dataset, we progressed through iterations of improvements. In every iteration, 100 random tweets were annotated by three annotators\footnote{They are linguistic experts and their age ranges between $20$-$35$ years.} following the initial set of annotation guidelines. The annotators resolved the ambiguous cases mutually. In successive iterations, we further addressed the unsettled tweets that necessitated clarifications in the annotation guidelines. We reconsidered all prior annotations for every change in the guideline to ensure that the annotations emulated the most advanced version of the annotation guidelines. The final sprint of pilot annotation included annotating another set of randomly chosen 100 tweets with the final guidelines. Following \citet{trautmann2020fine}, we calculated the inter-annotator agreement using the $\alpha_u$ agreement measure \cite{krippendorff2016reliability}. We computed the mean pairwise value per post, where each token can be being classified into two classes, claim, and non-claim. We obtained a more than satisfactory agreement score of 0.87. Finally, the entire Twitter dataset was annotated by the same annotators that carried out the prefatory pilot annotations. 

\subsubsection{General Instructions}
\begin{itemize}[leftmargin=*]

    \item A claim is a statement that says you strongly believe that something is true. The action of showing, using or stating something strongly. 
    \item We use tweets that are annotated with a binary label using LESA guidelines \cite{gupta-etal-2021-lesa}, which indicates whether a tweet is a claim or not. 
    \item The claim span is that part of a sentence that contains the semantic representation of the claim. 
        Example: @realDonaldTrump A lot of people are saying cocaine cures COVID-19. Claim span: cocaine cures COVID-19.
    \item Since our primary goal is to tackle misinformation in OSM, we majorly focus on claims that have some social impact. 
\end{itemize}

\subsubsection{Guidelines and Examples}
\label{sec:guideline-examples}
\begin{itemize}[leftmargin=*]
    \item In the case of facts, we annotate the fact/span that may not be known by everyone, for example, scientific facts or legal (law) facts, and doesn’t involve any commonsense. However, we do not include universal facts in the claim span. \\
    Example 1: “Water is colorless” is a universally known fact and hence should not be marked as claim span. \\
    Example 2: “Virus always mutate” is a scientific fact that may not be known by everyone. Hence we will annotate this fact as a claim. 
    
    \item An assertion about future eventualities/predictions will not be included in the claim span. Prediction is an extrapolation based on an assertion and is associated with a confidence level that can never be greater than or equal to 100\%. Thus, we will not consider predictions as a part of the claim span. \\
    Example: @realDonaldTrump Uh no actually The virus will never go away Scientists will develop a vaccine for it that should be ready by next June which will allow nearly everyone to be immune to \#CoronaVirus This really isn't hard to understand even for a very stable genius.
    
    \item A proverb is a simple, concrete, traditional saying that expresses a perceived truth based on common sense or experience which contains wisdom, truth, morals, and traditional views in a metaphorical, fixed, and memorizable form. The proverbs are not facts. The elements of proverbs should be annotated as claim span. \\
    Example: “Prevention is better than cure” is not a claim. 
    
    \item If a claim contains statistics or dates, they should be included in the span. But not all numbers are important. \\
    Example 1: “@FernandoSVZLA @AP So far ~50 people outside China have it with no deaths. If China was hiding information and it was more lethal, we would see that fairly quickly”. Here the claim span is [~50 people outside China have it with no deaths] \\
    Example 2: “57 round trip to LA thanks coronavirus”. The number is not important here. 

    \item In case there are multiple conclusive independent claims in one tweet, we annotate each one of them separately.  \\
    Example: “5 million left Wuhan before the lockdown. If they were really interested in knowing, they'd be testing at least 1 in 100 cases of all viral pneumonia.  They're limiting who's being tested so they aren't accused of lying.  Oh, and they might be asked to actually do something.” Claim span would consist of: [1] “5 million left Wuhan before the lockdown” and [2] “They're limiting who's being tested so they aren't accused of lying”

    \item Tweets that negate a possibly false claim are also considered to be claims. \\
    Example: “disinfectants are not a cure for coronavirus”. 
    
    \item Tweets ‘reporting’ something to be true or an instance to have happened or will happen are claims.

    \item In cases of claims made in the form of a conditional sentence, the premise/context would be included in the span. \\
    Examples: if you’ve been in the McDonald’s play place you’re immune to the coronavirus. 

    \item For claims containing humor/sarcasm, only the humorous phrase will be considered as a claim span if it has some social impact. For satire, the complete sentence will be considered. \\\
    Example: @TheRickWilson Drinking bleach and/or injecting Disinfectant will cure COVID19. And cancer, heart disease, OCD, schizophrenia and AIDS. And life. \#Covid19 \#COVID
    Claim: Drinking bleach and/or injecting Disinfectant will cure COVID19. And cancer, heart disease, OCD, schizophrenia and AIDS. And life.
  
    \item Personal experience will only be part of the claim phrase if they are opinions with societal impacts/implications. \\
    Example: Story about how \#HydroxyChloroquine likely help people recover from \#Coronavirus. IMO, it was never touted as the cure but as option for treatment doctors should consider and it appears to work in some cases....39 in one place. https://t.co/2hhi6aSVrY \\
    Claim: [it was never touted as the cure but as option for treatment doctors should consider and it appears to work in some cases....39 in one place.]

    \item A claim can be a sub-part of a question, only if it is not a direct question. \\
    Example: @FLOTUS Melania, do you approve of ingesting bleach and shining a bright light in the rectal area as a quick cure for \#COVID19 ? \#BeBest” \\
    Claim: [ingesting bleach and shining a bright light in the rectal area as a quick cure for \#COVID19]

    \item Ground/Reasoning to justify a claim will not be a part of the claim phrase. \\
    Example: Covid-19 vaccine development and deployment in China, when available, will be made a global public good, which will be China’s contribution to ensuring vaccine accessibility and affordability in developing countries \\
    Claim phrase: [Covid-19 vaccine development and deployment in China, when available, will be made a global public good]

    \item Mocking/attacking a group or individual is not a part of the claim phrase.\\
    Example: Because \#coronavirus has tremendous chances of getting cured but your anti-national agenda is worse than death \\
    Claim: [coronavirus has tremendous chances of getting cured]

    \item Claim phrases do not include the predicate part that does not contribute to it being a claim. \\
    Example: I firmly believe that [if they found a way to bottle the @andersoncooper giggle, it would cure the corona virus ]

\end{itemize}

\subsection{Data Preprocessing}

We employ \textit{NLTK}\footnote{\url{https://www.nltk.org/}} to tokenize the tweets. Each token in the tweet is \textit{BIO} \textit{(Begin-Inside-Outside)} encoded to generate the labels \citep{ramshaw1999text}. Tag \textit{`B'} indicates that the token is at the start of a span, tag \textit{`I'} indicates that the token is within the span, while tag \textit{`O'} denotes that the token is outside the span. As RoBERTa tokenizes each word into subwords \citep{liu2019roberta}, each subword is given the BIO tag as per their parent word. We eliminate tokens made of non-ASCII and special characters, as well as remove the URLs provided in the tweets. Finally, we split hashtag terms by underscore delimiter and over non-consecutive uppercase character. For instance, \textit{\#WuhanLab} splits into \textit{`Wuhan'} and \textit{`Lab'}.


\end{document}